\documentclass[journal]{IEEEtran}
\usepackage{amsmath}
\usepackage{mathtools}
\usepackage{amssymb}
\usepackage{float}

\ifCLASSINFOpdf
\else
\fi

\begin{document}
\title{REMAP: Multi-layer entropy-guided pooling of dense CNN features for image retrieval}

\author{Syed~Sameed~Husain,~\IEEEmembership{Member,~IEEE,}
        Miroslaw~Bober,~\IEEEmembership{Member,~IEEE}
        \thanks{S. Husain and M. Bober are with University of Surrey, Guildford, Surrey, UK.
E-mails: \{sameed.husain,m.bober\}@surrey.ac.uk }% <-this % stops a space
}

\markboth{Author's version of the article published in IEEE Trans. on Image Processing, 22 May 2019, \textcopyright 2019 IEEE, DOI: 10.1109/TIP.2019.2917234}{}

\maketitle

\begin{abstract}
This paper addresses the problem of very large-scale image retrieval, focusing on improving its accuracy and robustness. We target enhanced robustness of search to factors such as variations in illumination, object appearance and scale, partial occlusions, and cluttered backgrounds - particularly important when search is performed across very large datasets with significant variability.  
We propose a novel CNN-based global descriptor, called REMAP, which learns and aggregates a hierarchy of deep features from multiple CNN layers, and is trained end-to-end with a triplet loss.  REMAP explicitly learns discriminative features which are mutually-supportive and complementary at various semantic levels of visual abstraction. These dense local features are max-pooled spatially at each layer, within multi-scale overlapping regions, before aggregation into a single image-level descriptor. To identify the semantically useful regions and layers for retrieval, we propose to measure the information gain of each region and layer using KL-divergence. Our system effectively learns during training how useful various regions and layers are and weights them accordingly. We show that such relative entropy-guided aggregation outperforms classical CNN-based aggregation controlled by SGD. The entire framework is trained in an end-to-end fashion, outperforming the latest state-of-the-art results. On image retrieval datasets Holidays, Oxford and MPEG, the REMAP descriptor achieves mAP of 95.5\%, 91.5\% and 80.1\% respectively, outperforming any results published to date. REMAP also formed the core of the winning submission to the Google Landmark Retrieval Challenge on Kaggle.
\end{abstract}


\IEEEpeerreviewmaketitle

\section{Introduction}

Research in visual search has become one of the most popular directions in the area of pattern analysis and machine intelligence. With dramatic growth in the multimedia industry, the need for an effective and computationally efficient visual search engine has become increasingly important. Given a large corpus of images, the aim is to retrieve individual images depicting instances of a user-specified object, scene or location. Important applications include management of multimedia content, mobile commerce, surveillance, medical imaging, augmented reality, robotics, organization of personal photos and many more. Robust and accurate visual search is challenging due to factors such as changing object appearance, viewpoints and scale, partial occlusions, varying backgrounds and imaging conditions. Furthermore, today's systems must be scalable to billions of images due to the huge volumes of multimedia data available.

In order to overcome these challenges, a compact and discriminative image representation is required. Convolutional Neural Networks (CNNs) delivered effective solutions to many computer vision tasks, including image classification. However, they have yet to bring anticipated performance gains to the image retrieval problem, especially on very large scales. The main reason is that two fundamental problems still remain largely open: (1) how to best aggregate deep features extracted by a CNN network into compact and discriminative image-level representations, and (2) how to train the resultant CNN-aggregator architecture for image retrieval tasks. 

This paper addresses the aforementioned problems by proposing a novel region-based aggregation approach employing multi-layered deep features, and developing the associated architecture which is trainable in an end-to-end fashion. Our descriptor is called REMAP – for Region-Entropy\footnote{We will use the term {\em region entropy} to mean relative entropy between the distributions of distances for matching and non-matching image pairs, or their KL-divergence}  based Multi-layer Abstraction Pooling; the name reflecting the key innovations. 
Our key contributions include:
\begin{itemize}
	\item 
	we propose to aggregate a hierarchy of deep features from different CNN layers, representing various levels of visual abstraction, and -importantly- show how to train such a representation within an end-to-end framework,
	\item 
	we develop a novel approach to ensembling of multi-resolution region-based features, which explicitly employs regions discriminative power, measured by the respective Kullback-Leibler (KL) divergence \cite{Kullback59} values, to control the aggregation process,
	\item 
	we show that this relative entropy-guided aggregation outperforms conventional CNN-based aggregations: MAC \cite{Tolias2015}, NetVLAD \cite{NetVLAD}, Fisher Vector \cite{Perronnin10}, GEM \cite{GEM} and RMAC \cite{Gordo2017},
	\item
	we compare the performance of three state-of-the-art base CNN architectures VGG16 \cite{vgg16}, ResNet101\cite{resnet} and ResNeXt101 \cite{RESNEXT} when integrated with our novel REMAP representation and also against existing state-of-the-art models. 
\end{itemize}

The overall architecture consists of a baseline CNN (e.g. VGG or ResNet) followed by the REMAP network. The CNN component produces dense, deep convolutional features that are aggregated by our REMAP method. The CNN filter weights and REMAP parameters (for multiple local regions) are trained simultaneously, adapting to the evolving distributions of deep descriptors and optimizing the multi-region aggregation parameters throughout the course of training. The proposed contributions are fully complementary and result in a system that outperforms not only the latest state-of-the-art in global descriptors, but can also compete with systems employing re-ranking based on local features. The significant performance improvements are demonstrated in detailed experimental evaluation, which uses classical datasets (Holidays \cite{JegouHE}, Oxford \cite{Philbin07}) extended by the MPEG dataset \cite{MPEGCDVS} and with up-to 1M distractors. 

This paper is organized as follows. Related work is discussed in Section \ref{sec:related}. The REMAP novel components and the compact REMAP signature are presented in Section \ref{sec:REMAP}. Our extensive experimentation is described in Section \ref{sec:experiments}. Comparison with the state-of-the-art is presented in Section \ref{com} and finally conclusions are drawn in Section \ref{sec:conclusions}. 

\section{Related work}\label{sec:related}
This section reviews important methods that have contributed to image-retrieval task.

\subsection{Methods based on hand-crafted descriptors}
Early approaches typically involve extracting multiple local descriptors (usually hand-crafted), and combining them into a fixed length image-level representation for fast matching. Local descriptors may be scale-invariant and centered on image feature points, such as for SIFT \cite{Lowe04}, or extracted on regular, dense grids, possibly at multiple scales independently of the image content \cite{dense1}. An impressive number of local descriptors have been developed over years, each claiming superiority, making it difficult to select the best one for the job - an attempt at comparative study can be found in \cite{Simone}. It should be noted that descriptor dimension (for hand-crafted features) is typically between 32 and 192, which is an order or two less than the number of deep features available for each image region. 

Virtually all aggregation schemes rely on clustering in feature space, with varying degree of sophistication: Bag-of-Words (BOW) \cite{BOW}, Vector of Locally Aggregated Descriptors (VLAD) \cite{Jegou12PAMI}, Fisher Vector (FV) \cite{Perronnin10}, and Robust Visual Descriptor (RVD) \cite{HusainPAMI}. BOW is effectively a fixed length histogram with descriptors assigned to the closest visual word; VLAD additionally encodes the positions of local descriptors within each voronoi region by computing their residuals; the Fisher Vector (FV) aggregates local descriptors using the Fisher Kernel framework (second order statistics), and RVD combines rank-based multi-assignment with robust accumulation to reduce the impact of outliers. 

\begin{figure*}
	\centering
	\includegraphics[width=1.0\textwidth]{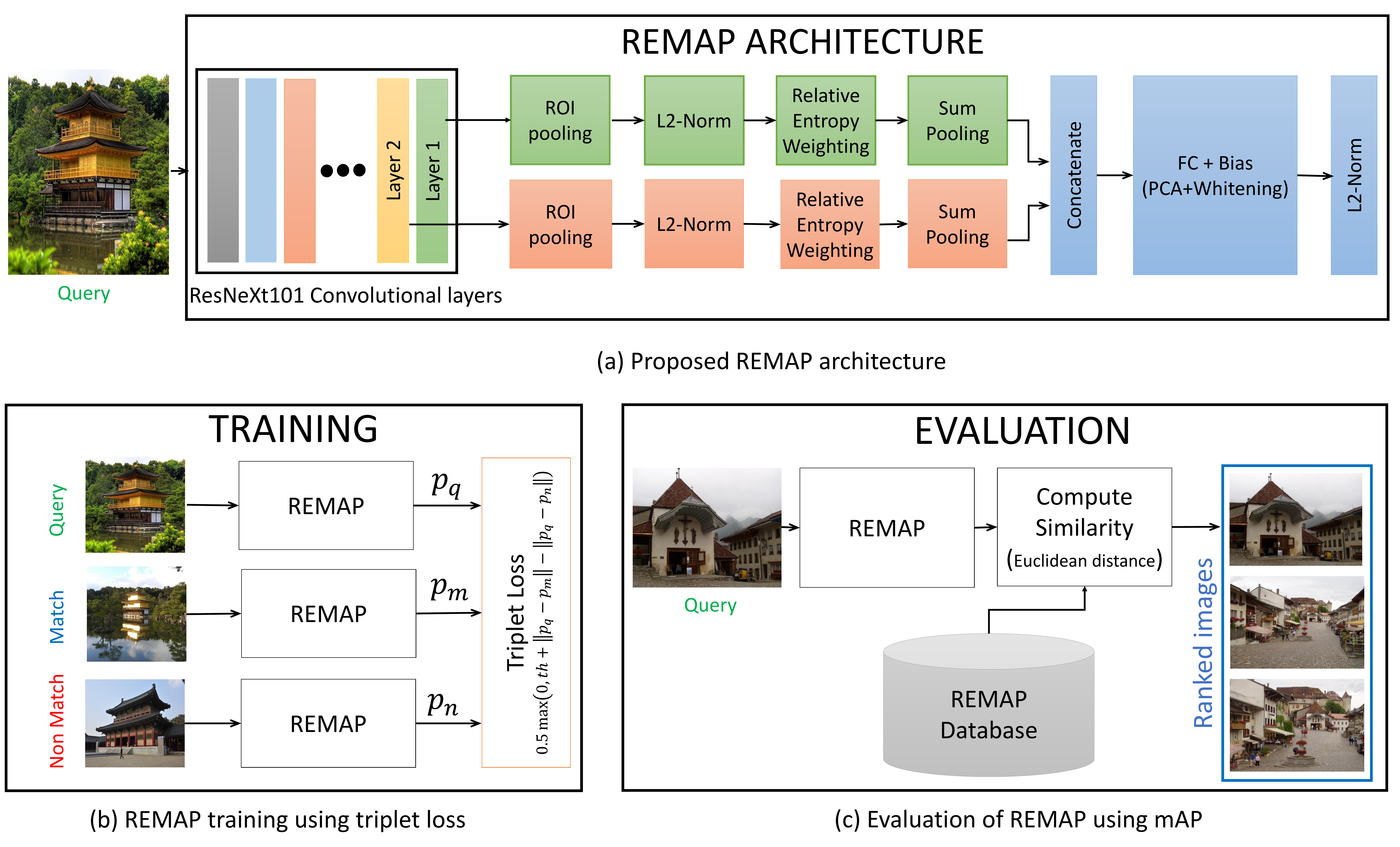}
	\caption{(a) Proposed REMAP architecture with KL-divergence based weighting (KLW) and Multi-layer aggregation (MLA). Note that Layer 1 is the last convolutional layer of ResNeXt101 architecture, (b) training of REMAP CNN using triplet loss on Landmarks dataset, (c) Evaluation of REMAP on state-of-the-art datasets}
	\label{RA}
\end{figure*}

\subsection{Methods based on CNN descriptors}
More recent approaches to image retrieval replace the low-level hand-crafted features with deep convolutional descriptors obtained from convolutional neural networks (CNNs), typically pre-trained on large-scale datasets such as the ImageNet. Azizpour et al. \cite{Razavian} compute an image-level representation by the max pooling aggregation of the last convolutional layer of VGGNet \cite{vgg16} and ALEXNET \cite{NIPS2012}. Babenko and Lempitsky \cite{SPOC} aggregated deep convolutional descriptors to form image signatures using Fisher Vectors (FV), Triangulation Embedding (TEMB) and  Sum-pooling of convolutional features (SPoC). Kalantidis et al. \cite{Kalantidi} extended this work by introducing cross-dimensional weighting in aggregation of CNN features. The retrieval performance is further improved when the RVD-W method is used for aggregation of CNN-based deep descriptors \cite{HusainPAMI}. Tolias et al. \cite{Tolias2015} proposed to extract Maximum Activations of Convolutions (MAC) descriptor from several multi-scale overlapping regions of the last convolutional layer feature map. The region-based descriptors are L2-normalized, Principal Component Analysis (PCA)+whitened \cite{JegouW}, L2-normalized again and finally sum-pooled to form a global signature called Regional Maximum  Activations  of  Convolutions  (RMAC). The RMAC dimensionality  is equal to the number of filters of last convolutional layer and is independent of the image resolution and the number of regions. In \cite{Seddati_2017}, Seddati et al. provided an in-depth study of several RMAC-based architectures and proposed a modified RMAC signature that combines multi-scale and two-layer feature extraction with feature selection. A detailed survey of content-based image retrieval (CBIR) methods based on hand-crafted and deep features is presented in \cite{7935507}.

\subsection{Methods based on fine-tuned CNN descriptors}

All of the aforementioned approaches use fixed pre-trained CNNs. However, these CNNs were trained for the purpose of image classification (e.g. 1000 classes of ImageNet), in a fashion blind to the aggregation method, and hence likely to perform sub-optimally in the task of image retrieval. To tackle this,  Radenovic et al. \cite{Radenovic2016CNNIR}, proposed to fine-tune MAC representation using the Flickr Landmarks dataset \cite{7299148}. More precisely, the MAC layer is added to the last convolutional layer of VGG or ResNet. The resultant network is then trained with a siamese architecture \cite{Radenovic2016CNNIR}, minimizing the contrastive loss. In \cite{GEM}, the MAC layer is replaced by trainable Generalized-Mean (GEM) pooling layer which significantly boosts retrieval accuracy. In \cite{Gordo2017}, Gordo et al. trained a siamese architecture with ranking loss to enhance the RMAC representation. The recent NetVLAD \cite{NetVLAD} consists of a standard CNN followed by a Vector of Locally Aggregated Descriptors (VLAD) layer that aggregates the last convolutional features into a fixed dimensional signature and its parameters are trainable via back-propagation. Ong et al. \cite{OngHB17} proposed SIAM-FV: an end-to-end architecture which aggregates deep descriptors using Fisher Vector Pooling. 

\begin{figure*}
	\centering
	\includegraphics[width=1.0\textwidth]{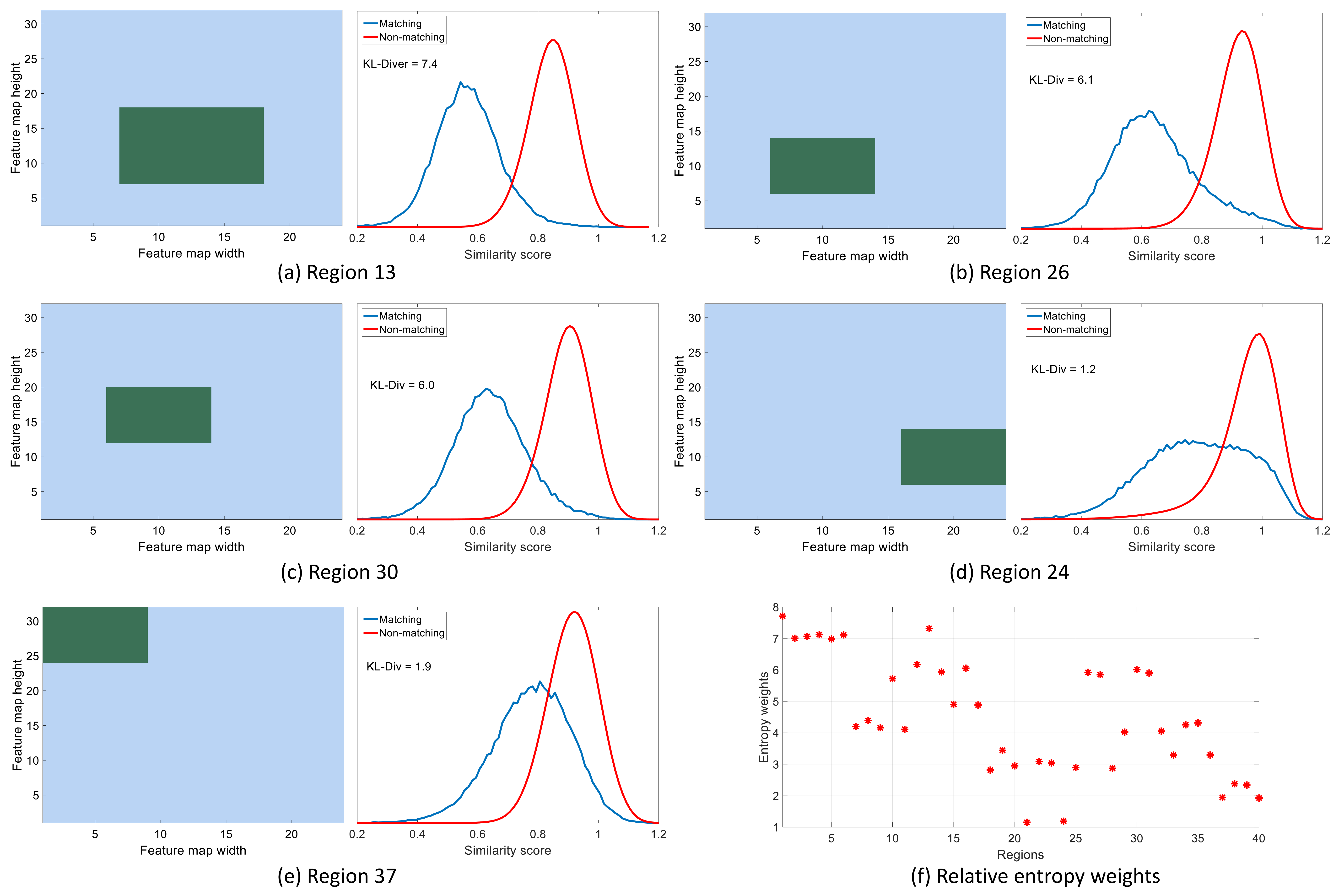}
	\caption{(a)-(e) Histogram of Euclidean similarity between matching and non-matching descriptors of different regions, (f) KL-divergence weights used to initialize the ROI weight vector $\alpha$ in REMAP }
	\label{KLW}
\end{figure*}

\section{REMAP representation}\label{sec:REMAP}

The design of our REMAP descriptor addresses two issues fundamental to solving content-based image retrieval: (i) a novel aggregation mechanism for multi-layer deep convolutional features extracted by a CNN network, and (ii) an advanced assembling of multi-region and multi-layer representations with end-to-end training.

The first novelty of our approach is to aggregate a hierarchy of deep features from different CNN layers, which are explicitly trained to represent multiple and complementary levels of visual feature abstraction, significantly enhancing recognition. Importantly, our multi-layer architecture is trained fully end-to-end and specifically for recognition. This means that multiple CNN layers are trained jointly to be: (1) {\em discriminative individually} (under the specific aggregation schemes employed within layers), (2) {\em complementary to each other} in recognition tasks, and (3) {\em supportive} to the extraction of the features required at subsequent layers. This contrasts with the MS-RMAC network \cite{Seddati_2017}, where no end-to-end training of the CNN is performed: fixed weights of the pre-trained CNN are used as a feature extractor. The important and novel component of our REMAP architecture is multi-layer end-to-end finetuning, where the CNN filter weights, relative entropy weights and PCA+Whitening weights are optimized simultaneously using Stochastic Gradient Descent (SGD) with the triplet loss function \cite{Gordo2017}. The end-to-end training of the CNN is critical, as it explicitly enforces intra-layer feature complementarity, significantly boosting performance. Without such joint multi-layer learning, the features from the additional layers - while coincidentally useful - are not-trained to be either discriminative nor complementary. The REMAP multi-layer processing can be seen in Figure \ref{RA}, where multiple parallel processing strands originate from the convolutional CNN layers, each including the ROI-pooling \cite{Tolias2015}, L2-normalization, relative entropy weighting and Sum-pooling, before being concatenated into a single descriptor.  

The region entropy weighting is another important innovation proposed in our approach. The idea is to estimate how discriminatory individual features are in each local region, and to use this knowledge to optimally control the subsequent sum-pooling operation. The region entropy is defined as the relative entropy between the distributions of distances for matching and non-matching image descriptor pairs, measured using the KL-divergence function \cite{Kullback59}. The regions which provide high separability (high KL-divergence) between matching and non-matching distributions are more informative in recognition and are therefore assigned higher weights. Thanks to our entropy-controlled pooling we can combine a denser set of region-based features, without the risk of less informative regions overwhelming the best contributors.
Practically, the KL-divergence Weighting (KLW) block in the REMAP architecture is implemented using a convolutional layer with weights initialized by the KL-divergence values and optimized using Stochastic Gradient Descent (SGD) on the triplet loss function.
 
The aggregated vectors are concatenated, PCA whitened and L2-normalized to form a global image descriptor. 

All blocks in the REMAP network represent differentiable operations therefore the entire architecture can be trained end-to-end. We perform training on the Landmarks-retrieval dataset using triplet loss - please see the Experimental Section for full details of the datasets and the training process. Additionally, the REMAP signatures for the test datasets are encoded using the Product Quantization (PQ) \cite{JegouPQ} approach to reduce the memory requirement and complexity of the retrieval system.

We will now describe in detail the components of the REMAP architecture, with reference to the Figure \ref{RA}. We can see that it comprises of a number of commonly used components, including the max-pool, sum-pool and L2-norm functions. We denote these functions as $Maxp(\mathbf{x}),Sump(\mathbf{x}), L2(\mathbf{x})$ respectively, where $\mathbf{x}$ represents an input tensor. 

We also employ the Region Of Interest (ROI) function \cite{Tolias2015}, $\zeta : \mathbb{R}^{w,h,d} \rightarrow \mathbb{R}^{r \times d}$. The ROI function $\zeta$ splits an input tensor of size $w\times h \times d$ into $r$ overlapping spatial blocks using a rigid grid and performs spatial max-pooling within regions, producing a single $d$-dimensional vector for each region. More precisely, the ROI block extracts square regions from CNN response map at $S$ different scales \cite{Tolias2015}. For each scale, the regions are extracted uniformly such that the overlap between consecutive regions is as close as possible to 40\%. The number of regions $r$ extracted by the ROI block depends on the image size $(1024\times 768 \times 3)$ and scale factor $S$. We performed experiments to determine the optimum number of regions for our REMAP network. It can be observed from Table \ref{region} that the best retrieval accuracy is obtained using $r$=40. This is consistent across all the experiments.
\begin{table}[h]
	\caption{Optimum number of regions $r$ for REMAP network}
	\label{region}
	\centering
	\begin{tabular}{|c|c|c|c|c|}
		\hline
	Scale ($S$)& Number of&  Holidays & Oxford & MPEG \\
	                & Regions ($r$) &(mAP)&(mAP)& (mAP) \\ \hline \hline
	2 & 8  & 91.3  & 72.4  & 63.2  \\ \hline
	3 & 20 & 92.4  & 74.2  & 67.4  \\ \hline
	\bf{4} & \bf{40} & \bf{93.3}  & \bf{76.1}  & \bf{69.3}  \\ \hline 
	5 & 70 & 93.0 & 75.6   & 68.5  \\ \hline
	\end{tabular}
\end{table}

\begin{figure*}
	\centering
	\includegraphics[width=1.0\textwidth]{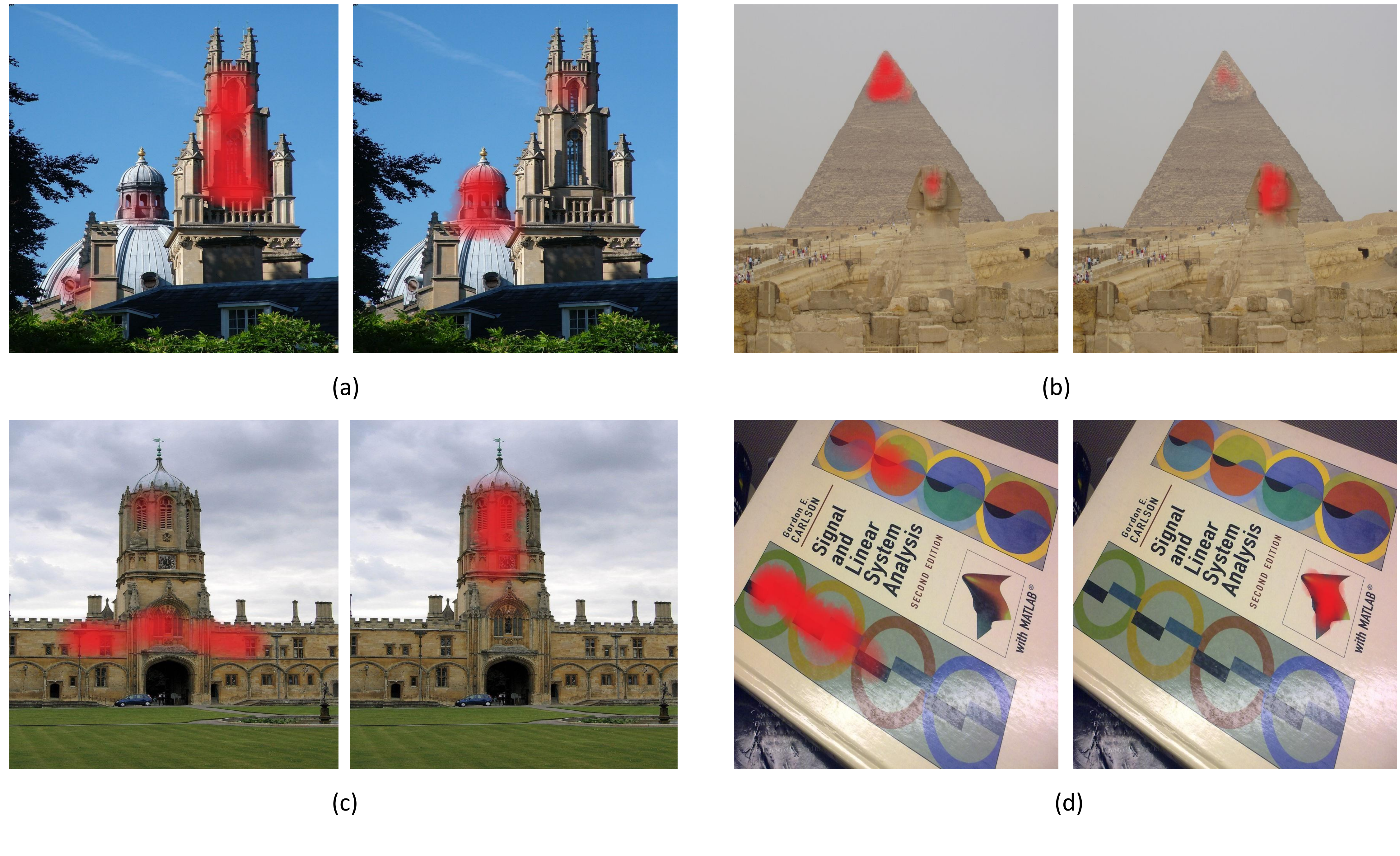}
	\caption{Activation maps for last and second last convolutional layers of an off-the-shelf ResNeXt101 feature extractor. It can be seen that the two layers focus on different but important features of the object thus justifying our multi-layer aggregation (MLA) approach. For each pair (a)-(d): left image shows the last layer, right image shows the second-last.}
	\label{Amaps}
\end{figure*}

\subsection{CNN Layer Access Function}
The base of the proposed REMAP architecture is formed by any of the existing CNN commonly used for retrieval, for example VGG16 \cite{vgg16}, ResNet101\cite{resnet} and ResNeXt101 \cite{RESNEXT}. All these CNNs are essentially a sequential composition of $L$ ``convolutional layers''. The exact nature of each of these blocks will differ between the CNNs. However, we can view each of these blocks as some function $l_i:\mathbb{R}^{w_i\times h_i \times d_i} \rightarrow \mathbb{R}^{w'_i\times h'_i \times d'_i} , 1\leq i \leq L$, that transforms its respective input tensor into some output tensor, where $w$, $h$ and $d$ denote the width, height and depth of the input tensor into a certain block and $w'$, $h'$ and $d'$ denote the height, width and depth of output tensor from that block.

The CNN can then be represented as the function composition: $f(\mathbf{x)} = l_L(l_{L-1}(...(l_1(\mathbf{x}))))$, where $\mathbf{x}$ is the input image of size $w_0 \times h_0$ with $d_0$ channels. For our purpose, we would like to  access the output of some intermediate convolutional layer. Therefore, we will create a ``layer access'' function: 
\begin{equation}
f_l(\mathbf{x}) = l_l(l_{l-1}(...(l_1(\mathbf{x}))))
\label{eq:cnn_layer_access}
\end{equation}
where $1 \leq l \leq L$. $f_l$ will output the convolutional output of layer $l$.

\subsection{Parallel Divergence-Guided ROI Streams}
The proposed REMAP architecture performs separate and distinct transformations on different CNN layer outputs via parallel divergence-guided ROI streams. Each stream takes as input the convolutional output of some CNN layer and performs ROI pooling on it. The output vectors of the ROI pooling are L2-normalized, weighted (based on their informativeness), and linearly combined to form a single aggregated representation.

Specifically, suppose we would like to use the output tensor of the layer $l'$ from the CNN as input for ROI processing. Now, let $\mathbf{o} = f_{l'}(\mathbf{x}), \mathbf{o} \in \mathbb{R}^{w,h,d}$ be the output tensor from the CNN's $l'$ convolutional layer given an input image $\mathbf{x}$. This is then given to the ROI block followed by L2 block, with the result denoted as: $\mathbf{r} = L2(\zeta(\mathbf{o}))$. The linear combination of the region vectors is then carried out by weighted sum:
\[
W(\mathbf{r}) = \sum^{r}_{i=1} \alpha_i \mathbf{r}(i)
\]
where $\mathbf{r}(i)$ denotes the $i^{th}$ column of matrix $\mathbf{r}$.

In summary, the ROI stream can be defined by the following function composition:
\[
P(\mathbf{x}; l', \mathbf{\alpha} ) = W( L2( \zeta( f_{l'}( \mathbf{x} ) )); \mathbf{\alpha} ) 
\]
where the set of linear combination weights is denoted as $\mathbf{\alpha} = \{\alpha_1, \alpha_2,...,\alpha_r \}$

In this work, the linear combination weights can be initialized differently, fixed as constants, or learnable in the SGD process. These in turn give rise to different existing CNN methods. In RMAC \cite{Gordo2017} architecture, the weights are fixed to 1 and not optimized during the end-to-end training stage: i.e. weight vector $\mathbf{\alpha} = \{1, 1,...,1\}$.

A drawback of the ROI-pooling method employed in RMAC is that it gives equal importance to all regional representations regardless of information content. We propose to measure the information gain of regions using the class-separability between the probability distributions of matching and non-matching descriptor pairs for each region.  Our algorithm to determine the relative entropy weights includes the following steps: (1) images of dimensionality $1024\times 768 \times 3$ are passed through the offline ResNeXt101 CNN, (2) the features from the ultimate convolution layers are then passed to the ROI block which splits an input tensor of size $32\times 24\times 2048$ into $40$ spatial blocks and performs spatial max-pooling within regions, producing a single $2048$-dimensional vector per region/layer, (3) for each region and each layer, we compute $Pr(y/m)$ and $Pr(y/n)$ as the probability density function of observing a Euclidean distance $y$ for a matching and non-matching descriptor pair respectively. KL-divergence measure is employed to compute the separability between matching and non-matching pdfs. 
It can be observed from Figure \ref{KLW} (a-e) that the KL-divergence value for different regions vary significantly. For example, region 13, 26 and 30 provides better separability (high KL-divergence) than region 24 and 37. 

We propose to assign learnable weights to regional descriptors before aggregation into REMAP to enhance the ability to focus on important regions in the image. Thus we view our CNN as an information-flow network, where we control the impact of various channels based on the observed information gain.  More precisely, the KL-divergence values for each region (Figure \ref{KLW}(f)) are used to initialize the ROI weight vector $a$. We enforce non-negativity on weight vector $a$ during the training process.

Practically, the KL-divergence weighting layer (KLW) is implemented using a convolutional operation with weights that can be learned using stochastic gradient descent on the triplet loss function. 

\subsection{Final REMAP Architecture}
We can now describe the proposed multi-stream REMAP. At the base is an existing Convolutional Neural Network (VGG or ResNet). The CNNs are essentially a sequential composition of $L$ ``convolutional layers'', $N$ of which are used in aggregation ($N<=L$). The output tensor of convolutional layer $l$ can be accessed using $f_l$ (Eq. \ref{eq:cnn_layer_access}). We denote the $N$ number of CNN layers that will be used in aggregation as: $\{l'_1,l'_2,...,l'_{N}\}$, where $l'_i \in \{l'_1,l'_2,...,l'_{L}\}$ for each $i = 1,2,...,N$. 

Associated with each of the above CNN layers $l \in \{l'_1,l'_2,...,l'_{N}\}$ is a set of ROI linear combination coefficients $\alpha_{l'_i} = \{\alpha_{l'_i,1},..., \alpha_{l'_i,r}\}$. As a result, we have $N$ parallel ROI streams, each with output $P(\mathbf{x}; l'_i, \mathbf{\alpha}_{l'_i} )$. The outputs of the $N$ ROI streams are concatenated together into a high-dimensional vector: $\mathbf{p} = [P(\mathbf{x}; l'_1, \mathbf{\alpha}_{l'_1}), ..., P(\mathbf{x}; l'_{N}, \mathbf{\alpha}_{l'_N}) ]^T$. We then pass $\mathbf{p}$ to a fully connected layer with weights initialized by PCA+Whitening coefficients \cite{Gordo2017}.

In Table \ref{MLP}, we perform experiments on Holidays, Oxford and MPEG  to demonstrate how different convolutional layers of off-the-shelf ResNeXt101 perform when combined within the REMAP architecture. It is interesting to note that, individually, the best retrieval accuracy on the Holidays and MPEG datasets is provided by layer 2, and not by the bottleneck layer 1. Layer 1 (the last convolutional layer) delivers best performance only on the Oxford dataset. 
%We then investigate the retrieval accuracy achieved by 
The performance of layer 3 is lowest since it is too sensitive to local deformation. However, the philosophy of our design is to combine different convolutional layers, so we investigate the performance of such combinations (shown in the lower half of the table). It can be observed from Table \ref{MLP} that multi-layer REMAP significantly outperforms any single-layer representation. In the final REMAP representation we use the combination of the last two convolutions layers (layer 1+2), which are trained jointly, as this provides the best balance between the retrieval accuracy and the computational complexity of the training process. 
In Figure \ref{Amaps}, we visualize the maximum activation responses of last two convolutional layers of off-the-shelf ResNeXt101. It can be seen that the two layers focuses on different but important features of the object thus justifying our multi-layer aggregation (MLA) approach.

\begin{table}[h]
	\caption{Performance of Multi-layer Pooling}
	\label{MLP}
	\centering
	\begin{tabular}{|c||c|c|c|}
		\hline
	Method&  Holidays & Oxford & MPEG \  \\ \hline \hline
	layer 1 & 92.1  & 73.5  & 67.4  \\ \hline
	layer 2 & 92.4  & 73.1  & 67.9 \\ \hline
	layer 3 & 91.1  & 72.1  & 65.0  \\ \hline \hline
	layer 1+2 & 93.3 & 76.1  & 69.3  \\ \hline
	layer 1+3 & 92.7 & 75.3  & 68.4  \\ \hline
	layer 1+2+3 & 93.4 & 76.3  & 69.5  \\ \hline
	\end{tabular}
\end{table}

\subsection{End-to-End Siamese learning for image retrieval}
An important feature of the REMAP architecture is that all its blocks are designed to represent differentiable operations. The fixed grid $ROI$ $pooling$ is differentiable \cite{RCNN}. Our novel component $KL$-$divergence$ $weighting$ $(KLW)$ can be implemented using 1D convolutional layer, with weights than can be optimized. The $Sum$-$pooling$ of regional descriptors, $L2$-$normalization$ and $Concatenation$ of multi-layer descriptors are also differentiable. The PCA+Whitening transformation can be implemented using a $Fully$-$connected$ $(FC)$ layer with bias. Therefore, we can learn the CNN filter weights and REMAP parameters (KLW weights and FC layer weights) simultaneously using SGD on the triplet loss function, adapting to the  evolving  distributions  of  deep  features  and optimizing the  multi-region  aggregation  parameters  over  the  course  of training.

We proceed by removing the last pooling layer, prediction layer and loss layer from ResNeXt101 (trained on ImageNet) and adding REMAP pipeline to the last two convolutional layers. We then adopt a three stream siamese architecture to finetune the REMAP network using triplet loss \cite{Gordo2017}. More precisely,  we are given a training dataset of $T$ triplets of images, each triplet consists of a query image, a matching image and a closest non-matching image (non-matching image with the most similar descriptor to query image descriptor). More precisely, let $p_q$ be a REMAP descriptor extracted from the query image, $p_m$ be a descriptor from the matching image, and $p_n$ be a descriptor from a non-matching image. The triplet loss can be computed as:
\begin{equation}
L=0.5 \; max(0,th + || p_q - p_m ||^2 -|| p_q - p_n ||^2),
\label{eq:loss}
\end{equation}
where $th$ parameter controls the margin of the classifier here i.e. the distance threshold parameter defining when the distance between matching and non-matching pairs is large enough not to be considered in the loss. The gradients with respect to loss $L$ are back-propagated through the three streams of the REMAP network, and the convolutional layers, KLW layer and PCA+whitening layer parameters get updated.

\subsection{Compact REMAP signature} \label{sec:CREMAP}
Encoding high-dimensional image representation as compact signature provides benefit in storage, extraction and matching speeds, especially for large scale image retrieval tasks. This section focuses on deriving a small footprint image descriptor from the core REMAP representation.
In the first approach, we pass an image thorough REMAP network to obtain $D$ dimensional descriptor and select the top $d$ dimensions out of $D$. 

The second approach is based on Product Quantization (PQ) algorithm \cite{Jegou12PAMI}, in which $D-dimensional$ REMAP descriptor is first split into $m$ sub-parts of equal length $D/m$. Each sub-part is quantized using a separate $K$-$means$ quantizer with $k=256$ cluster centres and encoded using $n=log_{2}(k)$ bits. The size of the PQ-embedded signature is $B=m\times n$ bits. At test time, the similarity between query vector and database vectors is computed using Asymmetric Distance Computation \cite{Jegou12PAMI}. 

\section{EXPERIMENTAL EVALUATION} \label{sec:experiments}
The purpose of this section is to evaluate the proposed REMAP architecture and compare it against latest state-of-the-art CNN architectures. We first present the experimental setup which includes the datasets and evaluation protocols. We then analyze, the impact of the novel components that constitute our method, namely KL-divergence based weighting of region descriptors and Multi-layer aggregation. Furthermore, we compare the retrieval performance of different feature aggregation method including MAC, RMAC, Fisher Vectors and REMAP on four varied datasets with up-to 1Million distractors. A comparison with the different CNN representations is presented at the end of this section.

\subsection{Training datasets}
We train on a subset of the Landmarks dataset used in the work of Babenko et al. \cite{BabenkoSCL14}, which contains approximately 100k images depicting 650 famous landmarks. It was collected through textual queries in the Yandex image search engine, and therefore contains a significant proportion of images unrelated to the landmarks, which we filter out and remove. Furthermore, to guarantee unbiased test results we exclude all images that overlap with the MPEG, Holidays and Oxford5k datasets used in testing. We call this subset the Landmarks-retrieval dataset. 

The process to remove images unrelated to the landmarks and to generate a list of matching image pairs for triplets generation is semi-automatic, and relies on local SIFT features detected with a Hessian-affine detector and aggregated with the RVDW descriptor \cite{HusainPAMI}. For each of the 650 landmark classes we manually select a query image, depicting a particular landmark, and compute its similarity (based on the RVDW global descriptors) to all remaining images in the same class. We then remove the images whose distance from query are greater than a certain threshold (outliers), forming the Landmarks-retrieval subset of 25k images.

To generate matching image pairs we randomly select fifty image pairs from each class in the Landmarks-retrieval dataset. RANSAC algorithm is applied to matching SIFT descriptors in order to filter out the pairs that are difficult to match (the number of inliers are less than 5  - extreme hard examples) or very easy to match (the number of inliers greater than 30 - extreme easy examples). This way, about 15k matching image pairs are selected for the finetuning based on the triplet loss function.

\subsection{Training configurations}
We use MATLAB toolbox MatConvNet \cite{VedaldiM} to perform training and evaluation. The state-of-the-art networks VGG16, ResNet101 and ResNeXt101 (all pre-trained on ImageNet) are downloaded in MATLAB format and Batch-normalization layers are merged into preceding convolutional layers for finetuning.

\subsection*{Finetuning with triplet loss}
Each aforementioned CNN is integrated with the REMAP network and the entire architecture is fine-tuned on Landmarks-retrieval dataset with triplet loss. The images are resized to $1024\times768$ pixels before passing through the network. Optimization is performed by the Stochastic Gradient Descent (SGD) algorithm with momentum $0.9$, learning rate of $10^{-3}$ and weight decay of $5\times 10^{-5}$. The triplet loss margin is set to 0.1. 

An important consideration during training process is the generation of triplets, as generating them randomly will yield triplets that incur no loss. To address this issue, we divide the 15k matching image pairs from Landmarks-retrieval dataset into 5 groups. The REMAP descriptors are extracted from 25k images using the current model. For each matching pair, the closest non-matching (hard negative) example is then chosen, forming a triplet, consisting of the following: query example; matching example; non-matching example. The hard negatives are remined once per group, i.e. after every 3000 triplets. 

Another consideration is the memory requirement during training, as the network trains with image size of 1024$\times$768 pixels and with three streams at same time. Finetuning with deep architectures, VGG16, ResNet101 and ResNeXt101, is memory consuming and we could only fit one triplet at a time on a $TITAN X$ GPU with 12 GB of memory. To make the training process effective, we update the model parameters after every 64 triplets. The training process takes approximately 3 days to complete.

\begin{table*}[t]
	\caption{Impact of KL-divergence based weighting (KLW)}
	\label{KLW_table}
	\centering
	\begin{tabular}{|c||c|c|c|c|c|c|c|}
		\hline
	Network& Initialization & Optimization    & Holidays & Oxford5k & MPEG  & Hol100k &Oxf105k   \  \\ \hline \hline
	RMAC      & $a=\{1, 1, ... , 1\}$  & No   & 93.7     &86.8      &75.1   &87.8     &83.7   \\ \hline
	RMAC+SGD  & $a=\{1, 1, ... , 1\}$  & SGD  & 93.9     &87.1      &75.2   &87.9     &84.0   \\ \hline
	KLW       & $a$=\{entropy-weights\}    & SGD  & 94.5     &88.3      &77.5   &89.4     &86.3  \\ \hline
	\end{tabular}
\end{table*}

\begin{figure}[!t]
	\centering
	\includegraphics[width=\columnwidth]{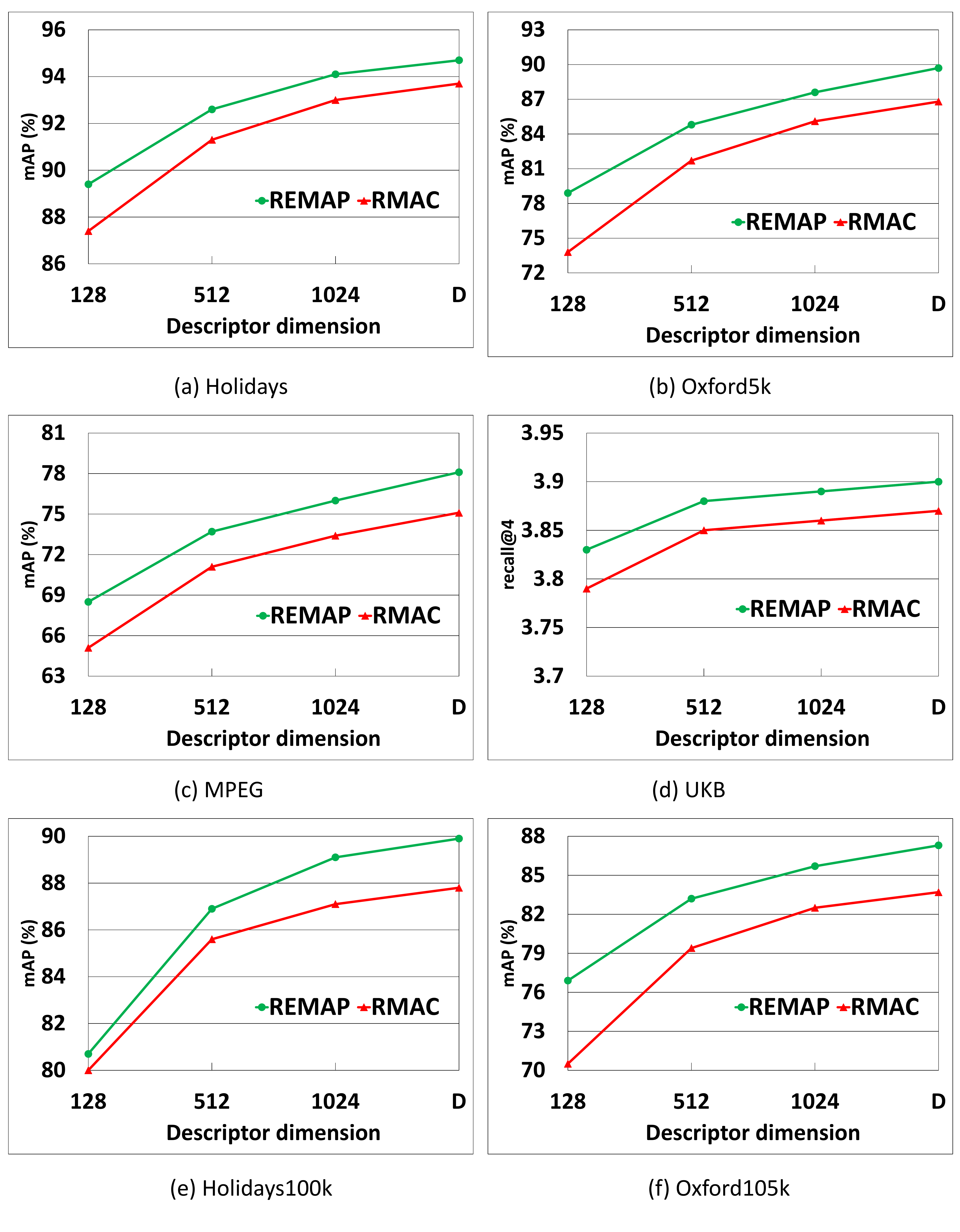}
	\caption{Impact of KL-divergence based weighting and Multi-layer aggregation (REMAP)
		(a) Holidays, (b) Oxford5k, (c) MPEG,  (d) UKB, (e) Holidays100k, (f) Oxford105k  (all results in mAP(\%) except for recall@4 for UKB);}
	\label{REMAP}
\end{figure}

\begin{figure}[!t]
	\centering
	\includegraphics[width=\columnwidth]{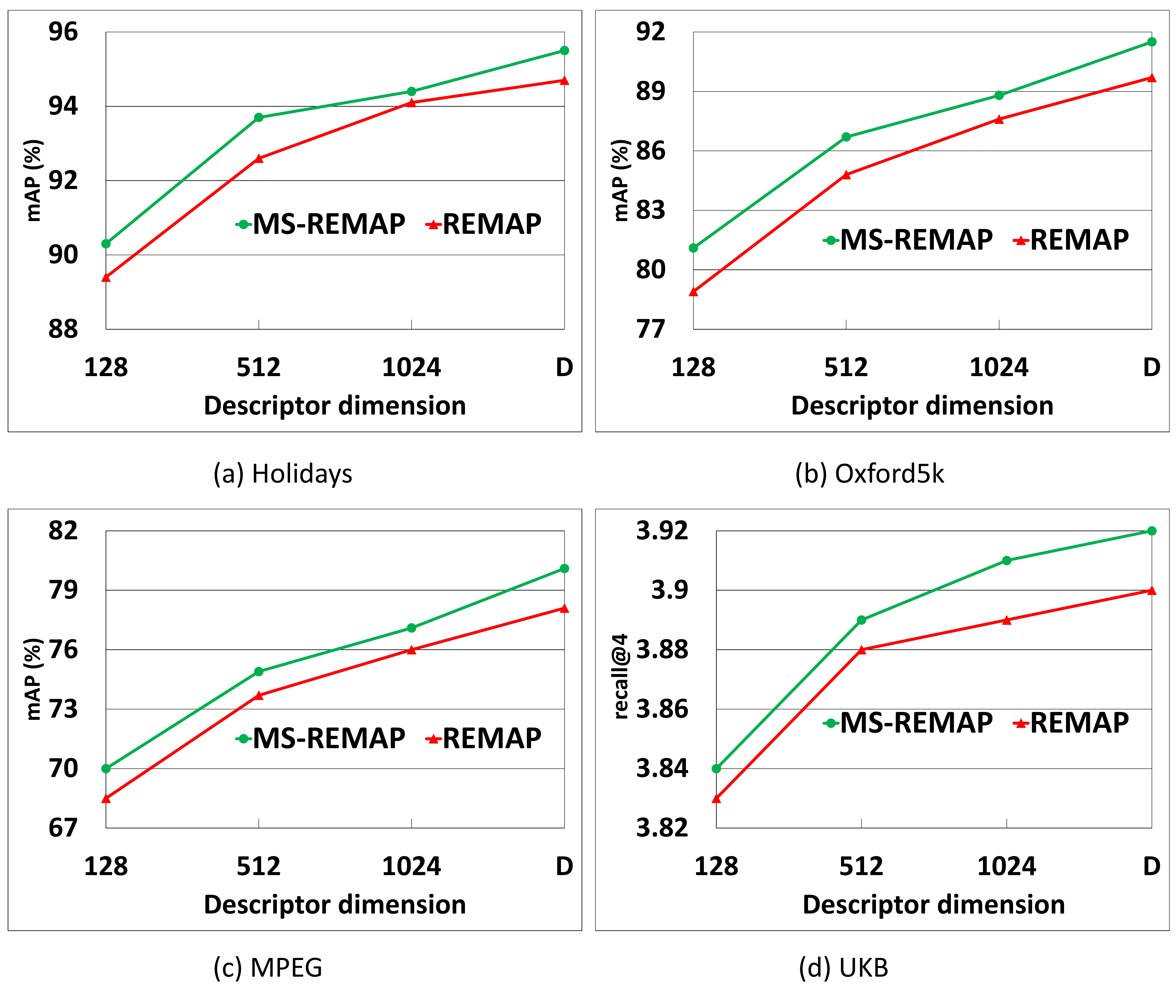}
	\caption{Multi-scale representation
		(a) Holidays, (b) Oxford5k, (c) MPEG,  (d) UKB  (all results in mAP(\%) except for recall@4 for UKB);}
	\label{MS-REMAP}
\end{figure}

\begin{figure*}
	\centering
	\includegraphics[width=\textwidth]{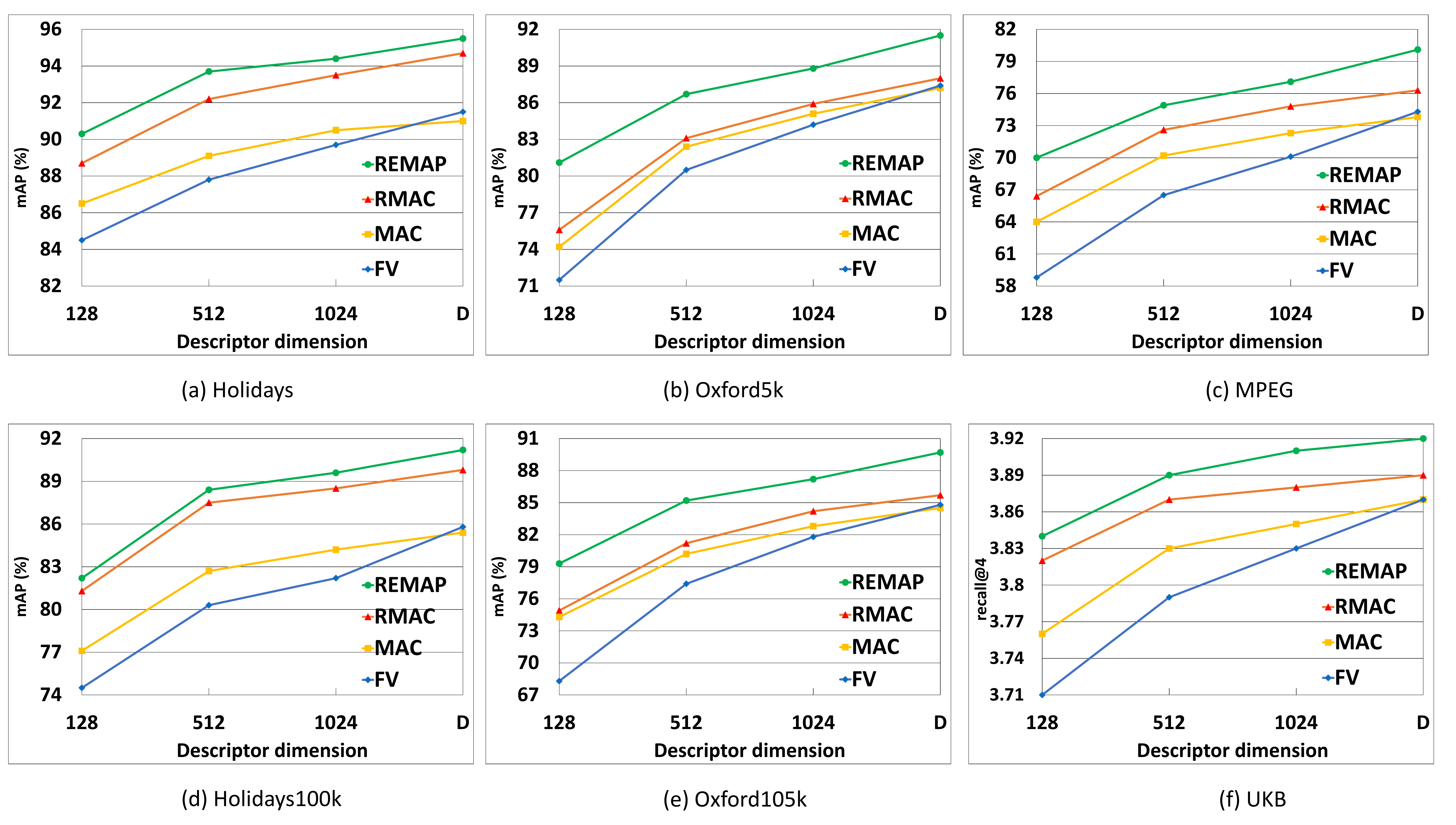}
	\caption{REMAP comparison with MAC, FV and RMAC
		(a) Holidays, (b) Oxford5k, (c) MPEG, (d) Holidays100k, (e) Oxford105k, (f) UKB  (all results in mAP(\%) except for recall@4 for UKB);}
	\label{REMAP_vs_RMAC}
\end{figure*}

\subsection{Test datasets}

The \textbf {INRIA Holidays} dataset \cite{JegouHE} contains 1491 holiday photos with a subset of 500 used as queries. Retrieval accuracy is measured by mean Average Precision (mAP), as defined in \cite{Philbin07}. To evaluate model retrieval accuracy in a more challenging scenario, the Holidays dataset is combined with 1 million distractor images obtained from Flickr, forming Holidays1M \cite{HusainPAMI}. 

The \textbf {University of Kentucky Benchmark} (UKB) \cite{Nister06} dataset comprises of 10200 images of 2550 objects. Here the performance measure is the average number of images returned in the first 4 positions (4 $\times$ Recall@4).

The \textbf {Oxford5k} dataset \cite{Philbin07} contains 5063 images of Oxford landmarks. The performance is evaluated using mAP over 55 queries. To test large scale retrieval, this dataset is augmented with 100k and 1 million Flickr images \cite{Bober13}, forming the Oxford105k \cite{Philbin07} and Oxford1M \cite{HusainPAMI} datasets respectively. We follow the state-of-the-art protocol for Oxford dataset and compute the image signature of query images using the cropped activations method \cite{NetVLAD} \cite{Radenovic2016CNNIR}.

The Motion Picture Experts Group (MPEG) have developed a heterogeneous and challenging \textbf {MPEG CDVS} dataset for evaluating the retrieval performance of image signatures \cite{MPEGCDVS}. The dataset contains 33590 images from five image categories (1) Graphics including Book, DVD covers, documents and business cards, (2) Photographs of Paintings, (3) Video frames, (4) Landmarks and (5) Common objects. A total of 8313 queries are used to evaluate the retrieval performance in terms of mAP.

The dimensionality of input images to the CNN is limited to 1024$\times$768 pixels.

\subsection{Impact of KL-divergence based weighting (KLW) and Multi-layer aggregation (MLA)}

In order to illustrate clearly and fairly the benefits of the novel elements proposed in our framework, we selected the best state-of-the art RMAC representation and integrated it with the latest ResNeXt101  architecture. We then performed fine-tuning, using procedures outline before. Please note that our fully optimized and finetuned RMAC representation outperforms results reported earlier on Oxford5k, MPEG and Oxford105k datasets and we use our improved result as the reference performance shown in the Table \ref{KLW_table} and Table \ref{MLA_table}. This represents the best state-of-the-art. We then introduce REMAP innovations, KL-divergence based weighting (KLW) and Multi-layer aggregation (MLA), and show the relative performance gain. Finally we combine all novel elements to show the overall improvement compared to baseline RMAC.

\subsubsection*{KL-divergence based weighting (KLW)}
We performed experiments to show that the initialization of the KLW block with relative entropy weights and then further optimization of weights using SGD is crucial to achieve optimum retrieval performance. We trained the following networks to compare:
\begin{itemize}
	\item The baseline is the RMAC representation in which the ROI weights are fixed ($a=\{1, 1, ... , 1\}$) and not optimized during the training process.
	\item In the second network RMAC+SGD, the weights are initialized with 1 ($a=\{1, 1, ... , 1\}$) and optimized using SGD on triplet loss function. 
	\item In the final network KLW, the relative entropy weights are initialized with KL-divergence values and further optimized using SGD on triplet loss.
\end{itemize}

It can be observed from Table \ref{KLW_table} that initialization of the KLW block with relative entropy weights is indeed significantly important for the network convergence and achieves best retrieval accuracy on all datasets. Furthermore, RMAC+SGD is not able to learn the optimum regional weights thus resulting in marginal improvement over RMAC. This is a very interesting result, which shows that optimization of the loss function alone may not always lead to optimal results and initialization (or optimization) of the network using the information gain may lead to improved performance. 

\subsubsection*{Multi-layer aggregation (MLA)}

Next, we perform experiments to show the advantage of Multi-layer aggregation of deep features (MLA). It can be observed from Table \ref{MLA_table} that MLA brings an improvement of +1.8\%, +2.6\% and +2.4\%  on Oxford5k, MPEG datasets and Oxford105k compared to single layer aggregation as employed in RMAC.

Finally, we combine the KLW and MLA blocks to form our novel REMAP signature and compare the retrieval accuracy with the RMAC reference signature, as a function of descriptor dimensionality. Figure \ref{REMAP} clearly demonstrates that REMAP significantly outperforms RMAC on all state-of-the-art datasets.

\begin{table}[h]
	\caption{Impact of Multi-layer aggregation (MLA) layer}
	\label{MLA_table}
	\centering
	\begin{tabular}{|c||c|c|c|c|c|}
    	\hline
    	Method& Holidays& Oxford5k& MPEG &Hol100k & Oxf105k   \\ \hline \hline
    	RMAC   &	93.7 & 	86.8  &	75.1   & 87.8  & 83.7            \\ \hline
    	MLA    &	94.3 &	88.6  &	77.7   & 88.9  & 86.1             \\ \hline
	\end{tabular}
\end{table}

\subsection{Multi-Scale REMAP (MS-REMAP)}
In this section, we evaluate the retrieval performance of MS-REMAP representation computed at test time without any further training. In MS-REMAP, the descriptors are extracted from images re-sized at two different scales and then aggregated into a single signature \cite{Gordo2017}. More precisely, let $X_1$ and $X_2$ be REMAP descriptors extracted from two images of sizes 1024$\times$768 and 1280$\times$960 pixels respectively. The MS-REMAP descriptor $X_m$ is computed by weighted aggregation of $X_1$ and $X_2$. 
\begin{equation}
    X_m=(2\times X_1) + (1.4\times X_2)
\label{eq:msr}
\end{equation}
It can be observed from Figure \ref{MS-REMAP} that multi-scale representation brings an average gain of 1\%, 1.8\% and 1.5\% on Holidays, Oxford5k and MPEG datasets compared to single scale representation. 

\subsection{Comparison of MAC, Fisher Vector, RMAC and REMAP networks}

In this section we compare the best network REMAP with state-of-the-art representations: MAC \cite{Tolias2015}, RMAC \cite{Gordo2017} and FV \cite{Perronnin10}. All the networks are trained end-to-end on Landmarks-retrieval dataset using triplet loss. We use Multi-Scale version for all representations.

In MAC pipeline, the MAX-pooling block is added to the last convolutional layer of the ResNeXt101. The MAX-pooling block is followed by PCA+Whitening and L2-Normalization blocks. The dimensonality of the output descriptor is 2048-D. 

For the Fisher Vector method, the last convolution layer is followed by Fisher Vector aggregation block, PCA+Whitening block and L2-Normalization block. 16 cluster centers are used for the Fisher Vector GMM model, with their parameters initialized using the EM algorithm. This resulted in FV of dimensionality 32k. The parameters of the CNN and Fisher vectors are trained using stochastic gradient descent on the triplet loss function.

In the RMAC pipeline, the last convolutional layers features are passed through rigid grid ROI-pooling block. The region based descriptors are L2-normalized, whitened with PCA and L2-normalized again. Finally, the normalized descriptors are aggregated using Sum-pooling block resulting in a 2048 dimensional signature.

Following conclusions can be drawn from the Figure \ref{REMAP_vs_RMAC}:
\begin{itemize}
	\item The RMAC representation outperforms MAC and Fisher Vector on all datasets.
	\item The full dimensional Fisher Vector signature achieves higher retrieval accuracy than MAC. However, FV suffers significantly from dimensionality reduction.
	\item REMAP signature on average outperforms all CNN representations. Compared to full dimensional RMAC, REMAP offers an gain of 1\%, 3.5\% and 3.8\% on Holidays, Oxford5k and MPEG dataset. The retrieval performance of REMAP is significantly better than RMAC
	(+1.6\%, 5.5\% and +3.6\% on Holidays, Oxford5k and MPEG) after the global descriptors are truncated to D=128.
\end{itemize}

\begin{figure}[!t]
	\centering
	\includegraphics[width=\columnwidth]{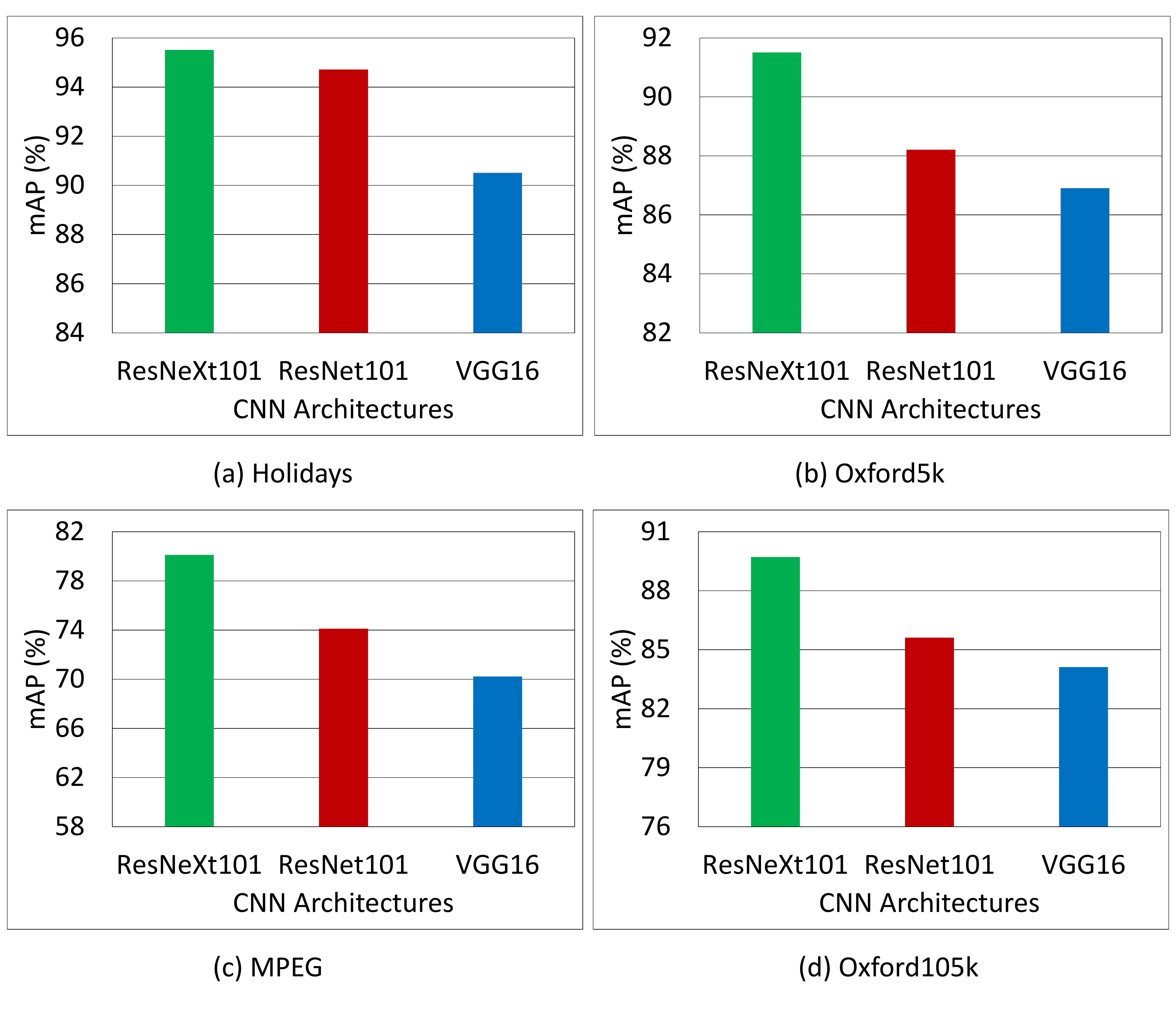}
	\caption{Comparison of CNN architectures
		(a) Holidays, (b) Oxford5k, (c) MPEG, (d) Oxford105k (all results in mAP(\%));}
	\label{Arch}
\end{figure}

\subsection{Convolutional Neural Network architectures}
In this section, we evaluate the performance of three state-of-the-art CNN architectures VGG16, ResNet101 and ResNeXt101 when combined with our REMAP network. All the networks are trained end-to-end on Landmarks-retrieval datatset. We use the Multi-Scale representation of REMAP to compare the CNNs. From the results shown in Figure \ref{Arch} we can observe that all three CNNs performed well on Holidays and Oxford dataset. The low performance on MPEG can be attributed to the fact that MPEG is a very diverse dataset (Graphics, Paintings, Videos, Landmarks and Objects) and our networks are finetuned only on landmarks. ResNeXt101 outperforms ResNet101 and VGG16 on all three datasets.

\subsection{Large scale experiments}
\begin{figure}[!t]
	\centering
	\includegraphics[width=\columnwidth]{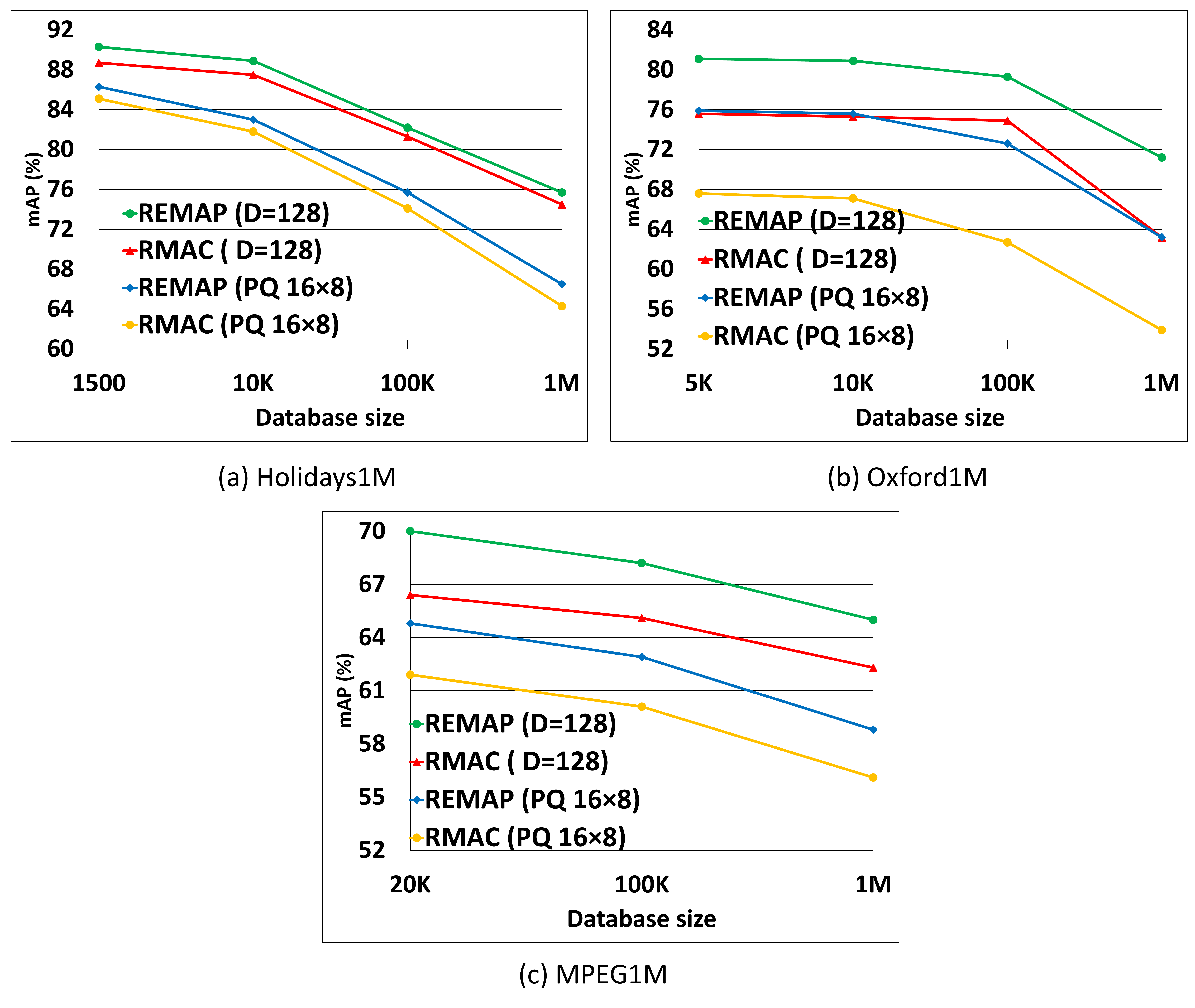}
	\caption{Retrieval performance as a function of database size
		(a) Holidays1M, (b) Oxford1M, (c) MPEG1M (all results in mAP(\%))}
	\label{LS}
\end{figure}

Figure \ref{LS} demonstrates the performance of our method on the large scale datasets of Holidays1M, Oxford1M and MPEG1M. The retrieval performance (mAP) is presented as a function of database size. We show the results for four methods:
\begin{itemize}
	\item the REMAP descriptor truncated to $D=128$;
	\item the RMAC descriptor truncated to $D=128$;
	\item the REMAP descriptor compressed to 16 bytes using $16\times8$ PQ;
	\item the RMAC descriptor compressed to 16 bytes;
\end{itemize}
The mAP performance clearly shows that REMAP signature outperforms RMAC on all datasets. On large scale datasets of Holidays1M, Oxford1M and MPEG1M, REMAP 128-D representation provides a gain +1.2\%, +8\% and +2.5\% over RMAC. Our very short image codes of 16 bytes achieves 66.5\%, 63.2\% and 58.8\% on Holidays1M, Oxford1M and MPEG1M, outperforming any results published to date.

 \begin{figure*}
	\centering
	\includegraphics[width=0.55\textwidth,height=3.7in]{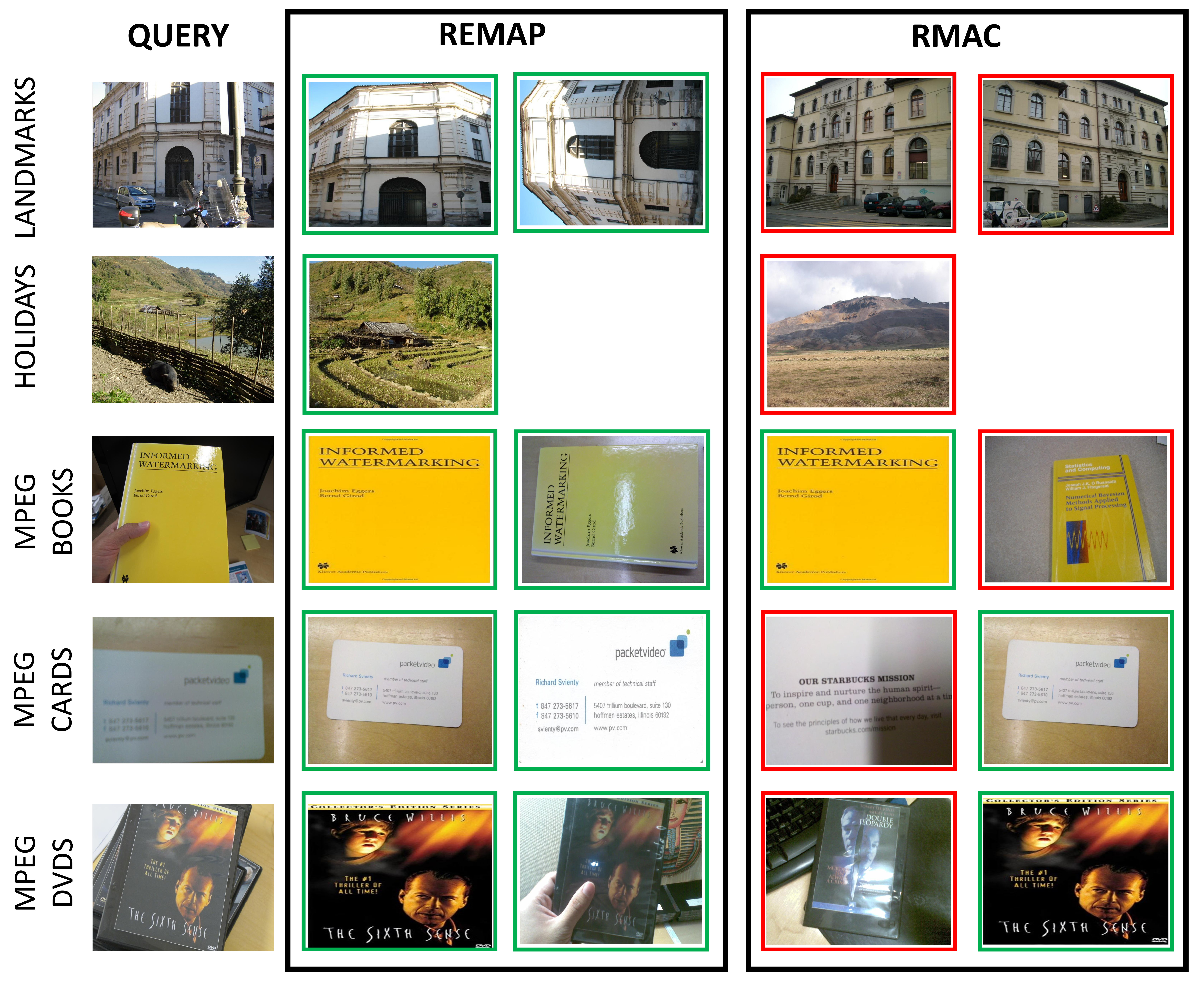}
	\caption{Example retrieval results for the REMAP and RMAC on the Holidays100k dataset and MPEG; images correctly retrieved are marked with green border.}
	\label{exam}
\end{figure*}

\section{Comparison with the state of the art} \label{com}

 This section compares the performance of REMAP with the latest state-of-the-art representations.
 
 Table \ref{tab:res8k} presents the performance for full dimensional representations. Practically the use of full dimensional descriptor is prohibitive due to search time and memory constraints; however the results are useful in determining the maximum capabilities of each representation.
 
 We categorize the global image representations into three categories: (1) representations derived from Hand-Crafted local features (HCF), (2) representations derived from Deep Convolutional features (DCF), (3) representations obtained from finetuned deep networks (FTF). 
 
 It can be observed that the proposed REMAP outperforms all prior state-of-the-art methods. Compared to the RMAC \cite{Gordo2017}, REMAP provides a significant improvement of +5\%, +1\% and +7.8\% in mAP on Oxford, Holidays and MPEG datasets. Furthermore, the REMAP signatures achieves a gain of +3.7\%, +1.6\% and  +6\% on Oxford, Holidays and MPEG datasets, over recently published GEM signature \cite{GEM}. The difference in retrieval accuracy, between REMAP and GEM, is even more significant on large scale datasets: Holidays100k (+3\%) and Oxford105k (+5\%). We also compare REMAP with our implementation ResNeXt+RMAC and the results show that REMAP representation is more robust and discriminative.
 
 REMAP also formed the core of the winning submission to the Google Landmark Retrieval Challenge on Kaggle \cite{landmarksolution18}. This gave us an opportunity to experimentally compare its performance on the Google landmark dataset \cite{landmarks-dataset18}. This new dataset is the largest worldwide dataset for image retrieval research, comprising more than a million images of 15K unique landmarks. More importantly, the evaluation was performed by Kaggle on a private unseen subset, preventing (unintentional) over-training. We evaluated REMAP, RMAC, MAC and SPoC aggregation applied to the ResNeXt network, without any additional modules (no query-expansion, DB-augmentation) - the results are shown in Table \ref{Google_table}. The REMAP architecture achieves mAP of $42.8\%$ and offers over 8\% gain over the closest competitor R-MAC. The classical SIFT-based system with geometric verification achieved only 12\%, illustrating clearly the gain brought by the CNN-based architectures. Our winning solution, which combined multiple signatures with query expansion (QE) and database augmentation (DA) achieved mAP of $62.7 \%$.
 
 \begin{table}[h]
	\caption{Performance on the Google Landmark Retrieval Challenge}
	\label{Google_table}
	\centering
	\begin{tabular}{|c||c|c|c|c|c|}
    	\hline
    	Method    & \bf{REMAP} & RMAC & MAC & SPoC & SIFT+ GV   \\ \hline \hline
    	mAP  &	\bf{42.8}  &	34.7  &	32.9   & 31.7  & 12             \\ \hline
	\end{tabular}
\end{table}

 It has recently become a standard technique to use Query Expansion (QE) \cite{Gordo2017}, \cite{GEM} to improve the retrieval accuracy. We applied QE to the REMAP
representation and it can be observed from Table \ref{tab:res8k} that REMAP+QE outperforms state-of-the-art results RMAC+QE \cite{Gordo2017} and GEM+QE \cite{GEM}.

 For the visualization purposes, Figure \ref{exam} shows 5 queries from Holidays100K and MPEG datasets where difference in recall between REMAP and RMAC is the biggest. We demonstrate the query and top ranked results obtained by REMAP and RMAC representations using these queries, where correct matches are shown by green frame.  
  \begin{table*}[t]
	\caption{Comparison with the state of the art using full dimensional signatures. HCF: Hand-Crafted Features, DCF: Deep Convolutional Features; FTF; Fine Tuned Features on landmarks datasets. The symbol $^*$ means that the images in Holidays dataset are manually rotated using an oracle. The symbol $^+$ means that the results are computed using the software provided by the authors.}
	\label{tab:res8k}
	\centering
	\begin{tabular}{|l|c|c|c|c|c|c|}\hline 
	Method & Features& Oxford5k & Oxford105k & Holidays & Holidays100k & MPEG \\ \hline \hline
	\multicolumn{7}{|c}{\bf Full dimensional descriptors} \\ \hline
	RVD-W \cite{HusainPAMI} &HCF  & 68.9  & 66.0    & 78.8     & -            & 65.9\\ \hline	
	TEMB \cite{Jegou14} &HCF   & 67.6  & 61.1    & 77.1     & -           & - \\ \hline	
	SPOC \cite{SPOC} &DCF   & 53.1  & 50.1    & 80.2$^*$   & -          &- \\ \hline	
	SPOC \cite{GEM} &DCF  & 68.1  & 61.1    & 83.9$^*$  & 75.1$^*$   &- \\ \hline	
	RVDW-CNN \cite{HusainPAMI} &DCF & 67.5 & 62.1   & 84.5     & 77.2     &  67.9 \\ \hline
	MAC \cite{GEM}  &DCF  & 56.4   & 47.8   & 79.0$^*$     & 66.1$^*$   &- \\ \hline
	RMAC \cite{Tolias2015} &DCF   & 66.9   & 61.6   & 86.9$^*$ & -           & - \\ \hline	
	RMAC (ResNet101) \cite{Gordo2017} &DCF  & 69.4 & 63.7 &91.3$^*$ & -           & - \\ \hline	
	CroW \cite{Kalantidi} &DCF  & 70.8  & 65.3    & 85.1   & -           & - \\ \hline	
    CWCF \cite{JimenezAN17} &DCF & 73.6 &  67.2       & - & -           & - \\ \hline
    MR-FS-CAM-RMAC  \cite{Seddati_2017} &DCF &72.3 &- &94.0$^*$ &- &- \\ \hline
	NetVLAD \cite{NetVLAD} &FTF & 71.6 & -       & 87.5$^*$ & -           & - \\ \hline
	RMAC \cite{Radenovic2016CNNIR} &FTF & 80.1 & 74.1 & 82.5$^*$ & 71.5$^*$ &- \\ \hline
	MAC \cite{Radenovic2016CNNIR} &FTF & 80.0 & 75.1  & 79.5$^*$ & 67.0$^*$ &- \\ \hline
	DELF \cite{Delf} &FTF &83.8 &82.6 &- &- &- \\ \hline 
	GEM \cite{GEM} &FTF & 87.8 & 84.6  & 93.9$^*$ &87.9$^*$ &74.0$^+$ \\ \hline
	RMAC \cite{Gordo2017} &FTF &86.1 & 82.8  & 90.3/94.8$^*$ & 88.9$^*$ &72.2$^+$ \\ \hline
	ResNeXt+RMAC &FTF &88.0 & 85.7  & 94.7$^*$ & 89.8$^*$ &76.3 \\ \hline
	REMAP &FTF &\bf{91.5} &\bf{89.7} &\bf{92.0/95.5$^*$} &\bf{91.2$^*$} &\bf{80.1} \\ \hline \hline
	\multicolumn{7}{|c}{\bf Query Expansion} \\ \hline
	RMAC+QE \cite{GEM} &FTF & 90.6 & 89.4  & - & -  &- \\ \hline
	GEM+QE \cite{GEM} &FTF & 91.0 & 89.5  & - & -  &- \\ \hline
    REMAP+QE &FTF &\bf{92.4} &\bf{91.4} &- &- &- \\ \hline
 	\end{tabular}
\end{table*}

\begin{table*}[h]
	\caption{Comparison with the state of the art using small footprint signatures. }
	\label{tab:res128k}
	\centering
	\begin{tabular}{|l|c|c|c|c|c|c|c|}\hline 
	Method & Dim & Oxford5k & Oxford105k &Oxford1M & Holidays & Holidays100k &Holidays1M  \\ \hline \hline
	SPOC \cite{HusainPAMI} &256 &58.9 &53.1 &41.1 &80.2$^*$ &73.9$^*$ &62.2$^*$ \\ \hline
	RVDW-CNN \cite{HusainPAMI} &256 &60.0 &55.9 &44.8 &81.3$^*$ &74.9$^*$ &63.5$^*$ \\ \hline
	NetVLAD \cite{NetVLAD} &256 &63.5 &60.8 &- &84.3$^*$ &- &- \\ \hline
	NetVLAD \cite{NetVLAD} &128 &61.4 &- &- &82.6$^*$ &- &- \\ \hline
	MAC \cite{Gordo2017} &128 &76.0 &68.2 &- &73.9$^*$ &- &- \\ \hline
	RMAC \cite{Gordo2017} &128 &73.0 &64.9 &- &80.0$^*$ &- &- \\ \hline
	RMAC+PCA \cite{Gordo2017}  &128 &77.9 &73.0 &62.1$^+$  &88.5$^*$ &80.5$^{*+}$ &72.7$^{*+}$ \\ \hline
	GEM \cite{GEM} &128 &79.5 &72.2 &61.0$^+$ &85.9$^*$ &76.5$^*$ &67.9$^{*+}$ \\ \hline
	REMAP  &128 &\bf{81.1} &\bf{79.3} &\bf{71.2} &\bf{90.3$^*$} &\bf{82.2$^*$} &\bf{75.7$^*$} \\ \hline
 	\end{tabular}
\end{table*}
 \subsection*{Compact image representation}
 This section focuses on a comparison of compact image signatures which are practicable in large-scale retrieval containing millions of images. To obtain a compact image descriptor, we pass an image through the trained REMAP network to obtain a 4096-dimensional descriptor. We then select top 128 dimensions out of 4096 dimensions The retrieval performance in Table \ref{tab:res128k} show that 128-D REMAP outperforms all state-of-the-art methods. The gain over GEM \cite{GEM} is +7\% and +5.7\% on Oxford105k and Holidays100k datasets. Compared to RMAC+PCA the gain is 6\% and 9\% on Oxford105k and Oxford1M datasets. On large scale datasets on Oxford1M and Holidays1M, REMAP significantly outperforms the best published results.
 
\section{Conclusion}\label{sec:conclusions}
In this paper we propose a novel CNN-based architecture, called REMAP, which learns a hierarchy of deep features representing different and complementary levels of visual abstraction. We aggregate a dense set of such multi-level CNN features, pooled within multiple spatial regions and combine them with weights reflecting their discriminative power. The weights are initialized by KL-divergence values for each spatial region and optimized end-to-end using SGD, jointly with the CNN features. The entire framework is trained in an end-to-end fashion using triplet loss, and extensive tests demonstrate that REMAP significantly outperforms the latest state-of-the art. 

\section*{Acknowledgements}
This work was partially supported by the Engineering and Physical Research Council (EPSRC) under InnovateUK grant 102811 (iTravel - A Virtual Journey Assistant) and by the UK Defence Science and Technology Laboratory (Dstl) and EPSRC under MURI grant EP/R018456/1.  The latter grant is part of the collaboration between US DOD, UK MOD and UK EPSRC under the Multidisciplinary University Research Initiative.

\ifCLASSOPTIONcaptionsoff
  \newpage
\fi

\bibliographystyle{IEEEtran}
\bibliography{REMAP}

\begin{IEEEbiography}[{\includegraphics[width=1in,height=1.2in]{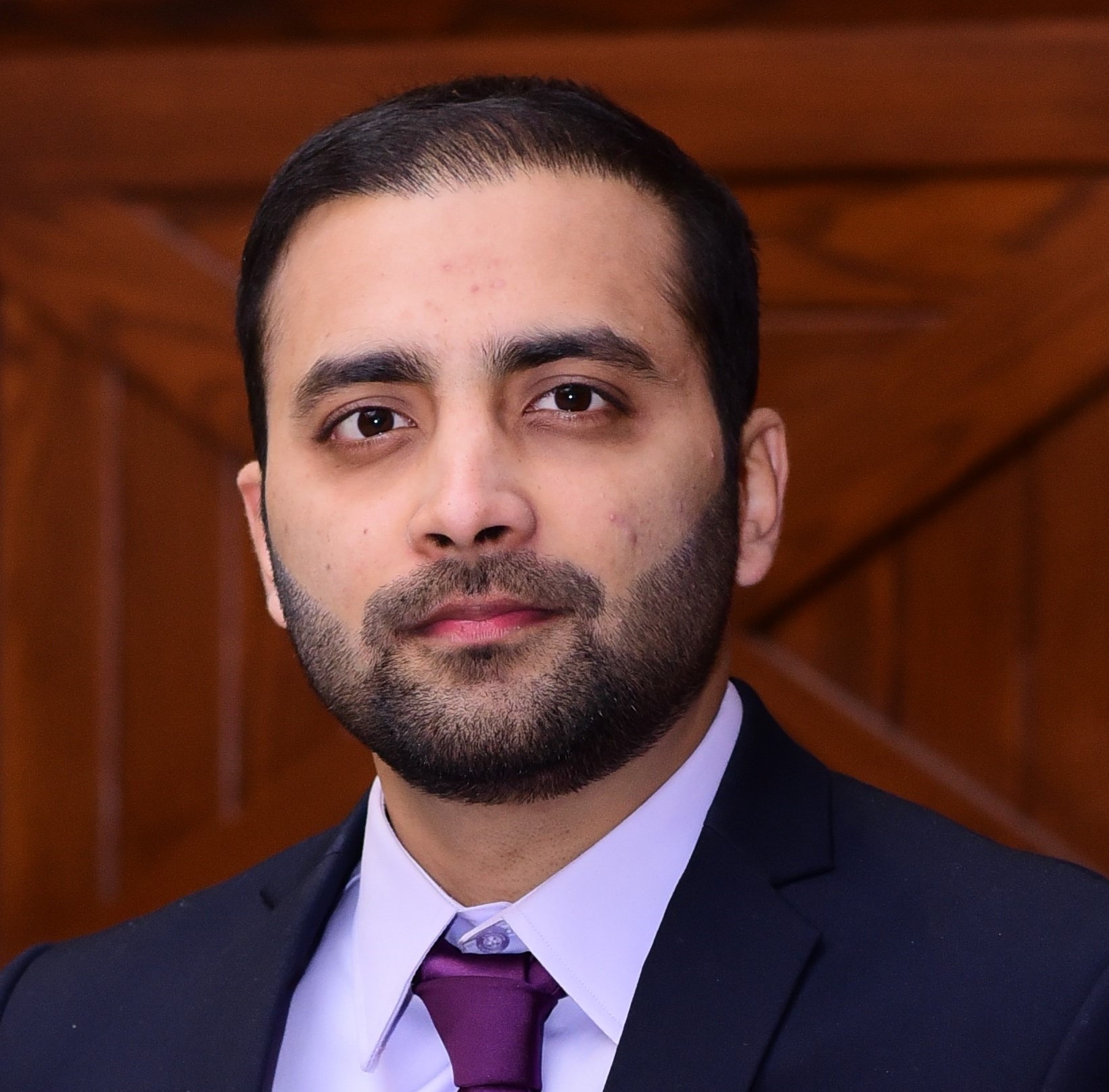}}]{Syed Sameed Husain} is a Research Fellow at the Centre for Vision, Speech and Signal Processing, University of Surrey, United Kingdom. He received MSc and PhD degrees from University of Surrey, in 2011 and 2016, respectively. His research interests include machine learning, computer vision, deep learning and image retrieval. Sameed's team has recently won the prestigious Google Landmark Retrieval Challenge. 

\end{IEEEbiography}
\begin{IEEEbiography}[{\includegraphics[width=1in,height=1.25in]{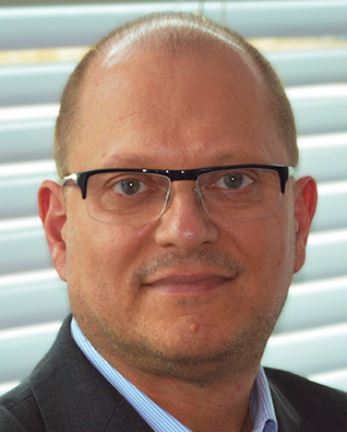}}]{Miroslaw Bober} is a Professor of Video Processing at the University of Surrey, U.K. In 2011 he co-founded Visual Atoms Ltd, a company specializing in visual analysis and search technologies. Between 1997 and 2011 he headed Mitsubishi Electric Corporate R\&D Center Europe (MERCE-UK). He received BSc degree from AGH University of Science and Technology, and MSc and PhD degrees from University of Surrey. His research interests include computer vision, machine learning and AI, with a focus on analysis and understanding of visual and multimodal data, and efficient representation of its semantic content. Miroslaw led the development of ISO MPEG standards for over 20 years, chairing the MPEG-7, CDVS and CVDA groups. He is an inventor of over 80 patents, many deployed in products. His publication record includes over 100 refereed publications, including three books and book chapters, and his visual search technologies recently won the Google Landmark Retrieval Challenge on Kaggle.
\end{IEEEbiography}

\end{document}